# FLOODDAMAGECAST: BUILDING FLOOD DAMAGE NOWCASTING WITH MACHINE LEARNING AND DATA AUGMENTATION

Chia-Fu Liu
Department of Civil and Environmental Engineering
Texas A&M University
College Station, TX 77843-3136
joeyliu0324@tamu.edu

Lipai Huang
Department of Civil and Environmental Engineering
Texas A&M University
College Station, TX 77843-3136
lipai.huang@tamu.edu

Kai Yin
Department of Civil and Environmental Engineering
Texas A&M University
College Station, TX 77843-3136
kai_yin@tamu.edu

Sam Brody
Department of Marine and Coastal Science
Texas A&M University at Galveston
Galveston, TX 77554-1675
brodys@tamu.edu

Ali Mostafavi
Department of Civil and Environmental Engineering
Texas A&M University
College Station, TX 77843-3136
amostafavi@civil.tamu.edu

## Abstract

Near-real time estimation of damage to buildings and infrastructure, referred to as damage nowcasting in this study, is crucial for empowering emergency responders to make informed decisions regarding evacuation orders and infrastructure repair priorities during disaster response and recovery. Here, we introduce FloodDamageCast, a machine learning (ML) framework tailored for property flood damage nowcasting. The framework leverages heterogeneous data to predict residential flood damage at a resolution of 500 meters by 500 meters within Harris County, Texas, during the 2017 Hurricane Harvey. To deal with data imbalance, FloodDamageCast incorporates a generative adversarial networks-based data augmentation coupled with an efficient machine learning model. The results demonstrate the framework's ability to identify high-damage spatial areas that would be overlooked by baseline models. Insights gleaned from flood damage nowcasting can assist emergency responders to more efficiently identify repair needs, allocate resources, and streamline on-the-ground inspections, thereby saving both time and effort.

## 1 Introduction

Flood hazards wreak havoc on urban areas, resulting in both physical destruction and loss of life in densely populated regions. In the United States alone, annual insurance claims have hovered around $1 billion per year over the past four decades [1]. This financial burden is expected to persist and potentially worsen due to the escalating frequency and intensity of flood events resulting from climate change [2, 3]. Rapid damage assessment of flooded areas is essential for swift response and recovery of affected communities. Emergency responders and public officials rely primarily on visual inspection to evaluate flood damage, incurring significantly delaying the recovery process. This limitation is due mainly to (1)

inadequately trained personnel conducting field inspections in the aftermath of flood events, and (2) limited flood data analytics capability available to emergency managers and public officials.

Expediting the flood damage assessment process is instrumental to accelerating post-disaster recovery efforts and bolstering community resilience against flood hazards, Currently, the main approach for estimating flood damage is based on specifying inundation depths then utilizing historical flood depth damage curves [4, 5]. The applicability of this approach for flood damage nowcasting, however, would be limited due to significant computation effort needed to model inundation depths using hydrological models based on the principles of hydrodynamics [6, 7, 8, 9]. Recent advancements in computing power and algorithms have rendered machine learning (ML) techniques increasingly prevalent and valuable across various flood risk analysis applications [10, 11, 12]. Given their versatility and adaptability, ML algorithms excel in integrating diverse datasets and discerning intricate relationships between input data and outputs. Consequently, ML approaches hold promise in mitigating the structure biases, data-specific requirements, and model calibration challenges commonly encountered in flood risk analysis.

In recent years, multiple studies have shown the capabilities of ML models for various flood risk analysis components, such as exposure mapping [13], early warning [14], and inundation nowcasting [15, 16]. Despite these advancements, the problem of reliable flood damage nowcasting at fine spatial resolution still persists with no existing model capable of producing satisfactory performance. This limitation is due mainly to three primary technical challenges. First, flood damage nowcasting would require consideration of features related to hydrological and topographic characteristics of areas, historical flood events, and the current flood characteristics, as well as built environment attributes. The consideration of various features and their non-linear interactions is a key step in flood damage nowcasting. Second, flood damage nowcasting involves dealing with highly imbalanced datasets. In a flood event in a region, a small percentage of buildings experience damage with the non-damage class being the prevalent class. Third, to build a model for damage nowcasting at fine resolution, a reliable ground truth is critical. In the U.S., the main data source for flood damage is the National Flood Insurance Program (NFIP) dataset. This dataset has two limitations: (1) the publicly available version is aggregated at the block group level and does not have the spatial resolution needed for damage nowcasting; (2) NFIP data does not include uninsured damage, which would negatively affect the realism of a model as residential buildings without insurance often get flooded.

To address these technical challenges, in this study, we introduce the FloodDamageCast model, an ML prediction framework tailored for near real-time residential property damage nowcasting based on built environment, topographical, and hydrological features and their interactions. The primary goal of FloodDamageCast is to expedite post-disaster property damage assessment to inform response and recovery efforts. To tackle the prevalent issue of class imbalance in flood risk classification tasks, our model integrates conditional tabular generative adversarial network (CTGAN) data augmentation techniques. An overview of the research components in creating FloodDamageCast is illustrated in Figure 1.

As shown in Figure 1., in creating FloodDamageCast, we use a light gradient-boosting machine (LightGBM) ML classification technique to assist in the nowcasting of flood building damage. LightGBM, a member of the gradient boosting decision tree (GBDT) algorithms, offers efficiency, accuracy, and interpretability, making it widely adopted in various ML tasks [17, 18]. Conventional GBDT algorithms face challenges in scanning all data instances for each feature to determine the optimal split point, leading to time-consuming operation. LightGBM addresses this issue with innovative techniques such as gradient-based one-side sampling (GOSS) and exclusive feature bunding (EFB). GOSS selectively samples instances with large gradients, enhancing efficiency, while EFB bundles exclusive features to reduce memory consumption. LightGBM has been widely applied across diverse fields owing to its high prediction accuracy, fast computational speed, and effectiveness in mitigating overfitting issues. Studies have showcased LightGBM's superiority over other GBDT methods in terms of learning performance, computational efficiency, and memory usage [19, 20, 21]. Its applications span various domains, including financial risk assessment [22, 23], transportation analytics [24, 25], disease prediction [26, 27, 28], social media analysis [29, 30, 31], hydrology, and flood prediction [32, 21, 33]. Notably, studies have demonstrated LightGBM's effectiveness in predicting flash flood susceptibility [21]; water levels in tide-affected estuaries [32]; and rapid urban flood prediction, surpassing results obtained from physical-based models [33]. These examples underscore LightGBM's versatility and effectiveness in various domains.

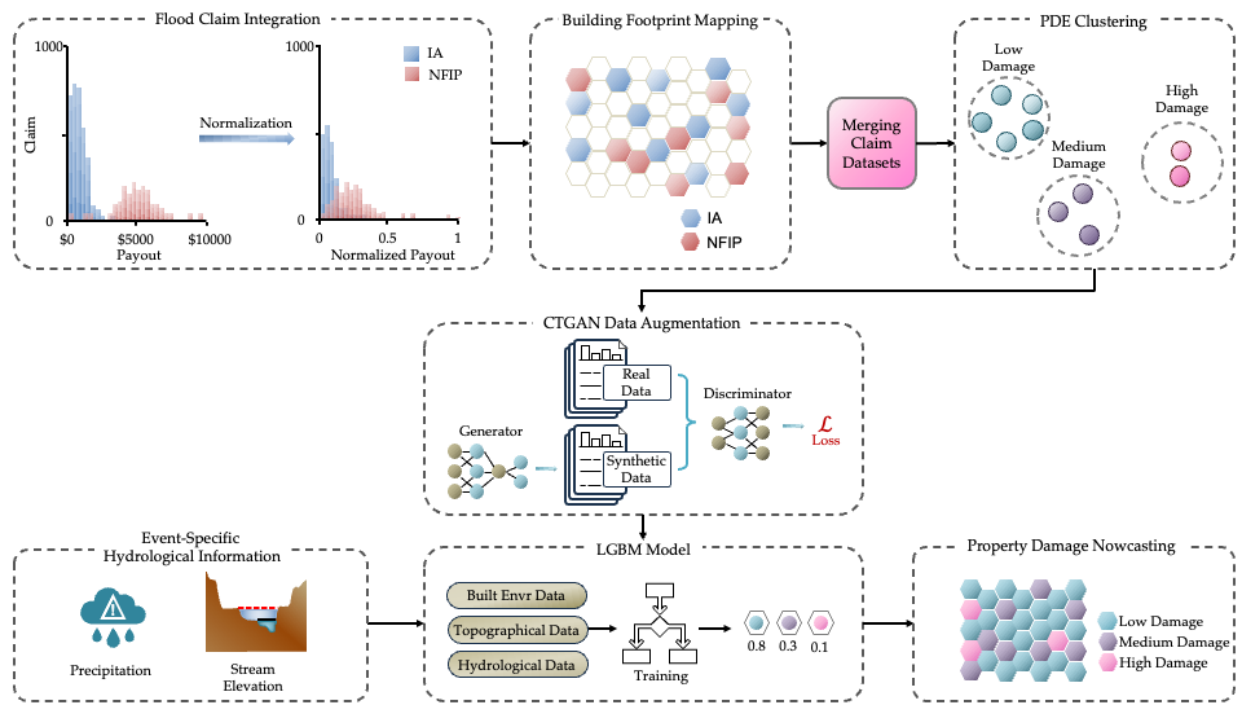

**Figure 1.** Schematic representation of the FloodDamageCast methodology

In classification of flood damages, algorithms often encounter the challenge of class imbalance in real-world datasets. These datasets frequently exhibit imbalanced distributions, with certain target values having significantly fewer observations than others. Specifically in the context of property damage classification, class imbalance arises when instances of high damage extent are notably rare compared to those of low damage extent, potentially biasing and compromising the performance of ML models. Previous studies have proposed various solutions—data-level, algorithms- level, and learning strategy methods—to address imbalanced dataset issues. Data-level methods typically involve techniques such as over-sampling the minority class, under-sampling the majority class, or a combination of both approaches [34, 35]. Algorithm-level methods, on the other hand, focus on alleviating the imbalanced class distribution during model training through cost-sensitive learning. This involves modifying the loss function to emphasize the minority classes [36, 37, 38]. Learning strategy methods aim to transform the imbalanced problem into a more balanced one by redefining the problem space or adjusting the learning strategy [39, 40]. CTGAN is specially designed for tabular data to effectively address class imbalances at both the data level and algorithm level [41]. CTGAN employs a training-by-sampling technique and a conditional generator that adeptly adjusts the simulated data distribution to favor minority classes. Hence, we use CTGAN for data augmentation in creating FloodDamageCast.

In sum, the novel aspects of this study are fivefold. First, the study presents FloodDamageCast as the first-of-its-kind ML model for flood damage nowcasting of residential buildings at fine spatial resolution. FloodDamageCast provides a rapid and automated tool for near-real-time damage estimation to enable flood nowcasting. Second, FloodDamageCast predicts residential flood damage extent based on a heterogeneous set of input features and their non-linear interactions. Hence, FloodDamageCast enables captured complex relationships among features that shape flood damage extent to provide a reliable estimation of flood damage. Third, FloodDamageCast includes a data augmentation component to address the issue of class imbalance in flood damage claim datasets to enhance the prediction performance of the model. Fourth, the use and integration of NFIP and IA claim datasets at fine-resolution in this study captures both insured and uninsured residential damage providing a reliable dataset for model training, testing and validation. Fifth, the explainability of FloodDamageCast clarifies features that shape residential flood damage extent to inform future flood mitigation and risk reduction plans and measures. These novel contributions inform interdisciplinary researchers in urban science, engineering, geography, and environmental science about new ML-based methods for implementing flood damage nowcasting and evaluating heterogeneous features and their importance in shaping flood damage extent. The model and findings also provide a novel data-driven tool for emergency managers and public officials to perform rapid and automated flood damage assessment as a flood event unfolds to inform their response and recovery efforts. In particular, FloodDamageCast enables expediting the damage assessment process that is now based mainly on on-the-ground inspections. Expediting damage assessments would facilitate faster processing of funding applications from the federal and state governments and accelerate the distribution of resources to impacted residents. FloodDamageCast enables prediction of residential flood damage extent based upon a heterogeneous set of input features obtained from publicly available datasets. Similar datasets exist in other regions and can be used in computing the features; thus, FloodDamageCast can be adapted by other regions based on local datasets to enable rapid and automated flood damage nowcasting. The structure of this work is as follows: Section 2 provides an overview of the study context and the data, including the target variable and the predictors employed in the FloodDamageCast model. In section 3, the methodology is detailed. Section 4 presents the model implementation, followed by the results in Section 5. Section 6 concludes the study.

## 2 Data

In this section, we introduce the target variable, property damage extent (PDE), which serves as an indicator of residential flood damage extent, along with the selection of predictors utilized in our FloodDamageCast model.

### 2.1 Flood property damage indicator

In August 2017, Hurricane Harvey, a Category 4 storm, made landfall along the Texas coast, bringing devastating impacts. The hurricane resulted in a rapid development of flooding due to record-breaking rainfall across much of the Harris County. This rainfall led to catastrophic drainage issues and significant rises in river/channel levels causing to residential properties. To assess the severity of residential property damage caused by Harvey flooding, this study uses flood claim data as a proxy for flood severity. These flood records, obtained from the National Flood Insurance Program (NFIP) and Individual Assistance (IA) program, offer valuable insights into the extent of flood-related damages.

NFIP offers subsidised flood insurance to property owners in participating communities, which map high-risk areas and regulate floodplain development [42]. Therefore, NFIP serves as a potentially significant source of disaster resources for flood-impacted households. However, property owners in non-participating communities are ineligible to purchase flood insurance. While NFIP insurance policies cover buildings and their contents, the IA program extends financial assistance to eligible individuals and households affected by disasters. IA is designed to address immediate needs of uninsured and underinsured disaster survivors, providing funds for lodging expense reimbursement, rental assistance, housing repair, personal property replacement [43]. It's important to note that the eligibility for IA is subject to an income screening process; families whose income surpasses 1.25 times the federal poverty line are referred to alternative loan programs [44]. Previous studies have focused primarily on NFIP records [45]; however, the lack of consideration of uninsured damages would adversely affect the completeness of the dataset for model training and testing. In our study, the consideration of IA dataset enables us to capture uninsured damages to enhance the completeness of the dataset used for model training and testing.

**Dataset processing.** We exclude damage to contents within buildings in our analysis, and instead focused on building structures. To present building damage value, we utilized building claim payments from NFIP and observed real property damage amounts assessed by FEMA in the IA programs. Records concerning damage unrelated to floof hazards under study in both datasets are ineligible to be paid by insurance. We included only instances with non-zero damage values from the NFIP and IA datasets to guarantee the accuracy of our analysis. Figure 2a displays the distribution of claim values for the two datasets.

**NFIP and IA integration.** To obtain a complete dataset for building damage extent, we integrated NFIP and IA claim sources using geographical analysis techniques. By mapping claims points to specific building polygons with high geographic resolutions [46], we ensure precise spatial alignment. For building polygons with NFIP records, only the NFIP claim record was utilized, disregarding the IA claim record due to their significantly smaller magnitude compared to NFIP claims. We included 41,606 instances from the NFIP dataset and 66,366 instances from the IA dataset in Harris County, Texas.

**Property damage extent.** As previously noted, NFIP and IA claim datasets vary in their magnitude. To obtain property damage extent as an indicator of residential flood damage, we used feature capping and min-max normalization techniques to scale the datasets appropriately before their amalgamation. The spatial distribution of the normalized claim value is illustrated in Figure 2c. Subsequently, we processed the normalized claim data into a 500-meter by 500-meter grid cell, enabling fine spatial representation of residential flood damage. In total, Harris County is segmented into 18,823 uniform and evenly-spaced grid cells. These grid cells were then categorized into three levels of property damage extent—class 0 (low damage), class 1 (medium damage), and class 2 (high damage)—using K-means clustering based on the summation of normalized data within each grid cell. Figure 2b illustrates the elbow method for determining the optimal value of k. The processed PDE values, indicative of the severity of building damage, serves as the target variable in our FloodDamageCast model. As depicted in Figure 2d, the PDE data exhibits significant imbalance, with the majority of spatial areas classified as low damage. Specifically, class 0, class 1, and class 2 comprise 18,149, 150, and 524 instances, respectively. This imbalance underscores the need for dealing data imbalance in order to achieve reliable model performance.

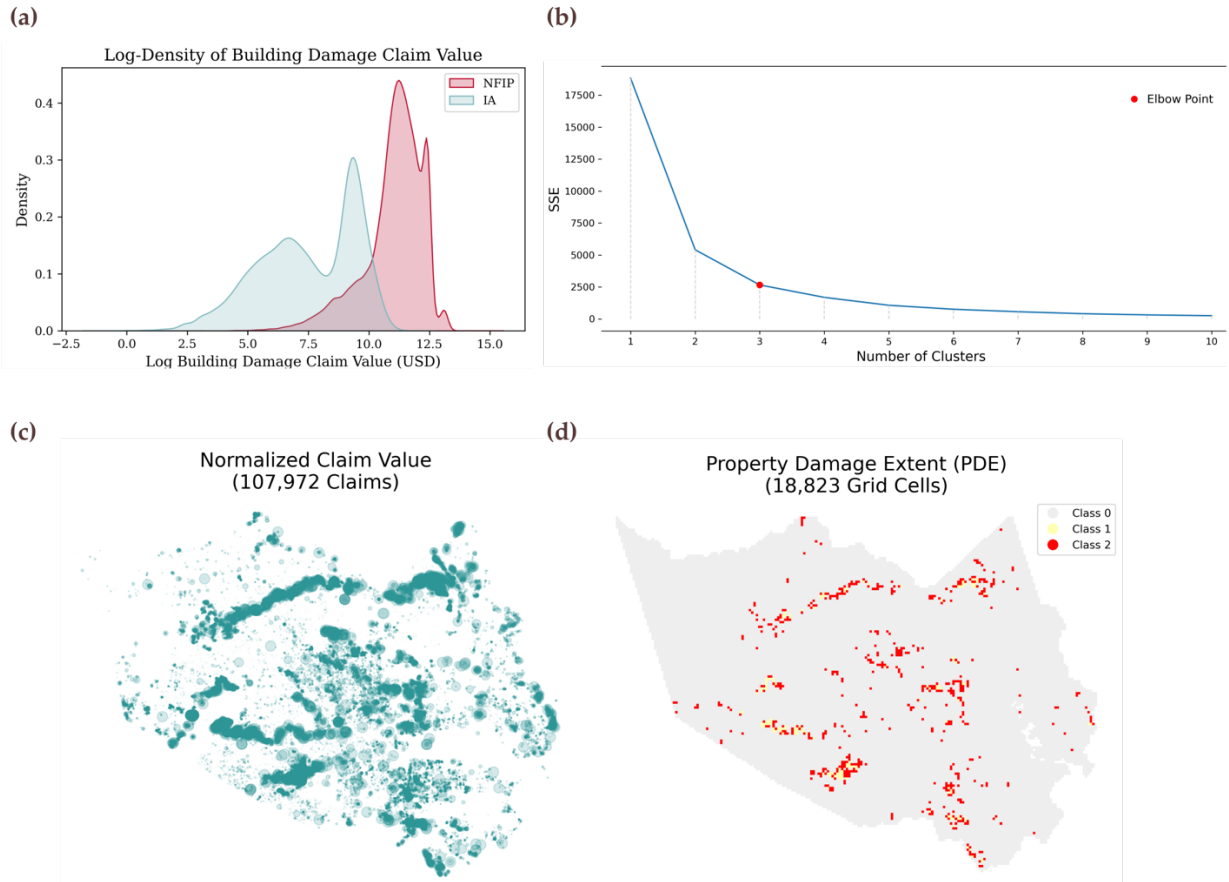

**Figure 2.** (a) Log-density plot of NFIP and IA claim value. (b) Determination of the optimal number of K-means clusters using the elbow method. (c) Spatial distribution of 107,972 normalized claim values. (d) Property damage extent (PDE) among 18,823 500-meter by 500-meter grid cells. The imbalance ratio of grid cells for class 0, class 1, and class 2 is 96.4%, 0.8%, and 2.8%, respectively.

## 2.2 Predictor selection

In this study, sixteen input features were carefully chosen based on review of the existing literature for determination of flood damage risk. While much of the data processing builds upon the previous work in flood risk prediction and community vulnerability assessment [47, 13, 16], this study refines the spatial resolution from 2-kilometer by 2-kilometer to 500-meter by 500-meter grid cells and incorporates two event-specific features for a more precise residential building damage nowcasting. These features fall into four distinct categories.

The first feature group encompasses aspects of built environment development, including building age, building area, building density, points of interest (POI) density, and population. These features encapsulate the intricate interplay of urban elements that significantly influence property damage post flooding. The second feature group delineates topographic features, including elevation, distance to coast, distance to stream, height above nearest drainage (HAND), imperviousness, and terrain roughness, providing insight into the terrain's physical characteristics. Third, hydrological data comprising historical flood frequency and flood level, which capture historical flood risk to inform flood damage nowcasting for an ongoing event. Last, rainfall data and stream status during Hurricane Harvey from August 25 to September 3, 2017, are included as event-specific hazard features, which could be updated in the model for future damage prediction. Rainfall data, representing the maximum daily rainfall amount, and stream status, obtained from 186 sensors in strategically placed gauge stations across Harris County bayous, provide crucial insights into flood dynamics during extreme weather events. Stream status categorizes stream elevation station statuses into non-flooding, flooding potential, or flooding likely, offering valuable information on flood risk levels. The data sources included in this study are described in Table 1.

**Table 1.** Description of predictors and data source. For numerical predictors, the mean (s.d.) is presented, while for categorical predictors—such as flood level and stream status—the mode is displayed.

| Category | Predictor | Statistics | Data Source |
| --- | --- | --- | --- |
| Built Environment Data | Building age | 29.1 (19.73) | National Structure Inventory dataset |
| | Building area | 2.6e+5 (3e.5) | National Structure Inventory dataset |
| | Building n umber | 77.9 (90.39) | Microsoft Building Footprints |
| | POI density | 5.4 (14.35) | Safegraph |
| | Foundation height | 0.6 (0.43) | National Structure Inventory dataset |
| | Population Number | 287.3 (351.22) | US Census Bureau dataset |
| Topographical Data | Elevation | 30.1 (17.53) | United States Geological Survey 3D Elevation dataset |
| | Distance to coast | 33.7 (22.4) | National Hydrography Dataset |
| | Distance to Stream | 6.5 (7.08) | National Hydrography Dataset |
| | Height above nearest drainage (HAND) | 7.2 (3.87) | National Flood Interoperability Experiment (NFIE) [48] |
| | Imperviousness | 36.5 (27.38) | National Land Cover Database |
| | Terrain roughness | 0.1 (0.09) | National Land Cover Database |
| Hydrological Data | Flood frequency | 4.46 (33.15) | FEMA historical flood claim dataset |
| | Flood level | 4 | FEMA historical flood claim dataset |
| Event-Specific Data | Rainfall | 16.3 (2.91) | Harris County Flood Warning System |
| | Stream status | 2 | Harris County Flood Warning System |

## 3 Methodology

### 3.1 GAN-based data augmentation

To address the imbalanced data issue in our study, we adopt data augmentation using generative adversarial networks (GAN). GAN, which has both a generator and a discriminator, has emerged as a powerful tool in synthetic data augmentation via a fusion neural network [49]. Through an adversarial training process, the generator produced synthetic data resembling the real data distribution while the discriminator distinguishes between real and synthetic data. These networks iteratively improve their performance until convergence. Conditional tabular generative adversarial network extends the GAN framework to handle structured data, such as tabular datasets, making it particularly adept at generating synthetic data for small, imbalanced datasets. By leveraging conditional generation capabilities, CTGAN can generate synthetic samples that closely mimic the statistical characteristics of the original data, even in scenarios with limited samples or imbalanced class distributions. In addition, CTGAN allows users to specify constraints or conditions during the data generation process, enabling the preservation of specific relationships or attributes present in the original dataset. This makes CTGAN a versatile and effective tool for augmenting tabular data, especially in scenarios where data scarcity or class imbalance poses challenges for traditional machine learning approaches.

Given the class imbalance in the flood damage data, along with the challenge of handling unseen data with potentially high variance, we augmented the existing dataset via CTGAN synthesizer [50]. In this step, we focused on generating more samples for the minority classes. To ensure similar performance between the synthetic dataset and the realistic dataset, we partitioned 80% of the original dataset for CTGAN data augmentation, reserving the remaining 20% as a test set for model evaluation. Critical hyperparameters, such as the learning rates for the generator and discriminator, and the number of training epochs for the CTGAN synthesizer, are optimized through grid search in the pretraining stage, which will be discussed later. To amplify the explainability of minority classes, we intentionally generated datasets of 50,000 records, allocating 60% for class 0, and 20% each for class 1 and class 2. This step ensures a balanced representation across the classes and enhances the model's ability to learn from underrepresented data.

### 3.2 Machine learning model

LightGBM is used to classify tasks in the FloodDamageCast. LightGBM has emerged as a leading-tree-based ML model, notable for its exceptional speed and satisfactory accuracy. Developed by Microsoft in 2017 [20] as a variant of the gradient boosting decision Tree (GBDT) algorithm, LightGBM incorporats several key improvements that distinguish it from traditional GBDT models. The first enhancement is the adoption of a histogram-based algorithm, which groups data into histograms with a predetermined number of bins. By simplifying the splitting point selection process based on these bins [51], LightGBM reduces computational complexity significantly compared to traditional GBDT models. Also, LightGBM employs leaf-wise tree growth instead of level-wise tree growth. Unlike level-wise growth, which progresses symmetrically, leaf-wise growth prioritized nodes with the greatest potential for error reduction. This approach accelerates the training process, especially with large datasets, by focusing on informative nodes. To prevent overfitting from deep tree growth, a maximum leaf-wise depth is specified. Moreover, LightGBM introduces GOSS (gradient-based one-side sampling) method to enhance subsampling efficiency. GOSS prioritizes instances with large gradients, which contribute significantly to tree construction, while randomly sampling instances with small gradients. This approach ensures efficient use of training data while mitigating overfitting. Another optimization technique employed by LightGBM is EFB (exclusive feature bunding), which groups features of high-dimensional data into a sparse feature space. This technique reduces computational burden by avoiding redundant feature calculations and maintaining high performance by preserving information effectively. These enhancements make LightGBM faster and more efficient than conventional GBDT models while preserving high performance.

As an ensemble method, LightGBM integrates the predictions from multiple decision trees (DT) to yield a generalized final prediction. The training process involves constructing a model with multiple trees, where each subsequent tree learns to predict the residuals of the preceding models. This additive training process is guided by minimizing a defined objective function, encompassing both training loss and regularization terms. To illustrate the training process of LightGBM for predicting PDE, consider a scenario in which a LightGBM model with T trees is constructed. Initially, the training dataset D = {($x_i$, $y_i$)}$_{i=1}^{n}$ is prepared, where $x_i \in R^m$ represents the m-dimensional input vector and $y_i \in Z$ denotes the one-dimensional categorical target PDE class value. The structure of LightGBM for PDE prediction is depicted in Figure 3.

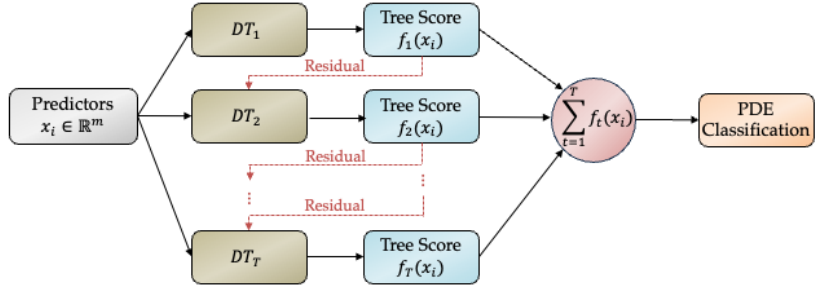

**Figure 3.** The structure of LightGBM for the PDE classification task.

In sum, the additive training process is:

$$\hat{y}_i = \sum_{t=1}^{T} f_t(x_i), f_t \in \mathcal{T}$$

In this case, $\hat{y}_i$ represents the prediction of $i^{th}$ example, $\mathcal{T}$ encompasses all possible tree structures, and $f_t$ denotes the learned function for the $t^{th}$ decision tree, obtained by minimizing the following objective:

$$f_t = \arg\min_{f_t} \sum_{i=1}^{n} L(y_i, \hat{y}_i^{(t)}) + \Omega(f_t)$$

where $L$ is the training loss function and $\Omega$ is the regularization function.

### 3.3 Performance metric

In this study, mean average precision (mAP) serves as the chosen performance metric to evaluate the learning algorithm. In a binary classification problem, where labels are either positive or negative, classification performance is often represented using a confusion matrix, as illustrated in Figure 4.

|  | | Predicted Label | |
|---|---|---|---|
| **Confusion Matrix** | | Positive | Negative |
| **Actual Label** | Positive | $T_p$ (True Positive) | $F_p$ (False Negative) |
| | Negative | $F_N$ (False Positive) | $T_N$ (True Negative) |

**Figure 4.** Confusion matrix for binary classification performance evaluation.

From these categories, precision (P) and recall (R) can be derived using the following formulas:

$$P = \frac{T_p}{T_p + F_p}$$

$$R = \frac{T_p}{T_p + F_N}$$

Examining the precision and recall formulas reveals a trade-off between their performances by adjusting the threshold for predicting an instance as a positive case. Unlike precision and recall, which are single-value metrics, average precision (AP) provides a comprehensive evaluation of a model's performance across multiple thresholds. AP is commonly used in information retrieval and binary classification tasks [52, 53], defined as:

$$AP = \sum_i (R_i - R_{i-1})P_i$$

where, $P_i$ and $R_i$ represent the precision and recall at the $i^{th}$ threshold, with each pair ($R_i$, $P_i$) referred to as an operating point. The calculation of AP involves computing precision at each threshold, weighted by the change in recall from the previous threshold, offering a nuanced understanding of the model's performance.

The precision-recall (PR) curve, which plots precision against the recall, provides a better measure compared to the receiver operating characteristic (ROC) curve, especially in imbalanced datasets. The PR is more sensitive to the class imbalance [54, 55], as it evaluates the precision (positive predictive value) against recall (true positive rate), focusing explicitly on the classifier's performance on the positive class. This sensitivity is crucial when the minority class is of interest, providing an accurate representation of the classifier's ability to identify instances of the minority class while minimizing false positives.

To extend AP to the multi-class scenario, mean average precision (mAP), is calculated by taking the average of AP across all the classes [56], offering a point of comparison between different multi-class classification problems:

$$mAP = \frac{1}{k}\sum_j^k AP_j$$

where, $AP_j$ represents the AP in the $j^{th}$ class, and $k$ denotes the total number of classes being evaluated. mAP provides a comprehensive evaluation of the classifier's performance across all classes, facilitating comparison between different multi-class classification tasks.

### 3.4 Pre-training

In the FloodDamageCast model, all numerical predictors related to built environment, topographical, and hydrological characteristics undergo min-max normalization to yield a more stable and robust learning process. Random under- sampling was then performed on the three PDE classes of the optimal CTGAN-augmented dataset to determine the optimal imbalance ratio for achieving the best performance.

Three major LightGBM hyperparameters were selected for optimization: (1) the number of leaves, (2) maximum tree depth, and (3) minimum data in a leaf. The number of leaves controls model complexity, with a larger number of leaf potentially resulting in overfitting. While theoretically, setting the number of leaves to 2 to the power of maximum depth might seem appropriate, the reality is more nuanced due to the deeper leaf-wise tree structure in LightGBM. Therefore, tuning the number of leaves alongside maximum depth is appropriate. Similarly, the minimum data in a leaf serves to prevent overfitting by setting a threshold for leaf creation. Its optimal value is contingent upon the number of leaves and the training set size. Consequently, a simultaneous optimization approach considering the interdependency of these hyperparameters was adopted. Hyperparameter optimization is conducted via random search, exploring the imbalance ratio for each class and the three major LightGBM hyperparameters. The search space for each hyperparameter is detailed in Table 2.

**Table 2.** Search space for random search hyperparameter optimization.

| Hyperparameter | Search space |
| --- | --- |
| Class 0 | [1000, 24000] |
| Class 1 | [1000, 8000] |
| Class 2 | [1000, 8000] |
| Number of leaves | [10, 50] |
| Maximum depth of the tree | [3, 15] |
| Minimum data in a leaf | [20, 100] |

**4 Model implementation**

The original dataset comprised 12,316 realistic records, characterized by a biased distribution among PDE classes, with percentages of 96.4% in class 0; 0.8% in class 1; and 2.8% in class 2. To address this class imbalance and augment the dataset, 80% of the realistic data was augmented using CTGAN. This process resulted in the generation of 120 datasets. Within each synthetic dataset, the ratio of classes 0, 1, and 2 is maintained at 3:1:1, effectively addressing the class imbalance present in the realistic dataset. These datasets were produced using CTGAN with various hyperparameter sets explored through grid search. Each augmented dataset consisted of 50,000 synthetic records prepared for pretraining. The optimal CTGAN configuration, yielding the highest mAP score of 0.834, was achieved with a generator learning rate of 0.0001, a discriminator learning rate of 0.0001, and training stoppage at 850 epochs. The pretraining result is shown in Figure 5a.

We performed a 1,000-iteration random search on the imbalance ratio of three classes and three model hyperparameters to train the FloodDamageCast model. The optimal imbalance ratio for the three classes was [23454, 6246, 7440] while the optimal number of leaves, maximum depth of the tree, and minimum data in a leaf were 41, 9, and 51 respectively. We also ran the LightGBM model on the original dataset without implementing CTGAN data augmentation to serve as the baseline model. The confusion matrix and the performance metric comparison are shown in Figure 5b and Figure 5c respectively.

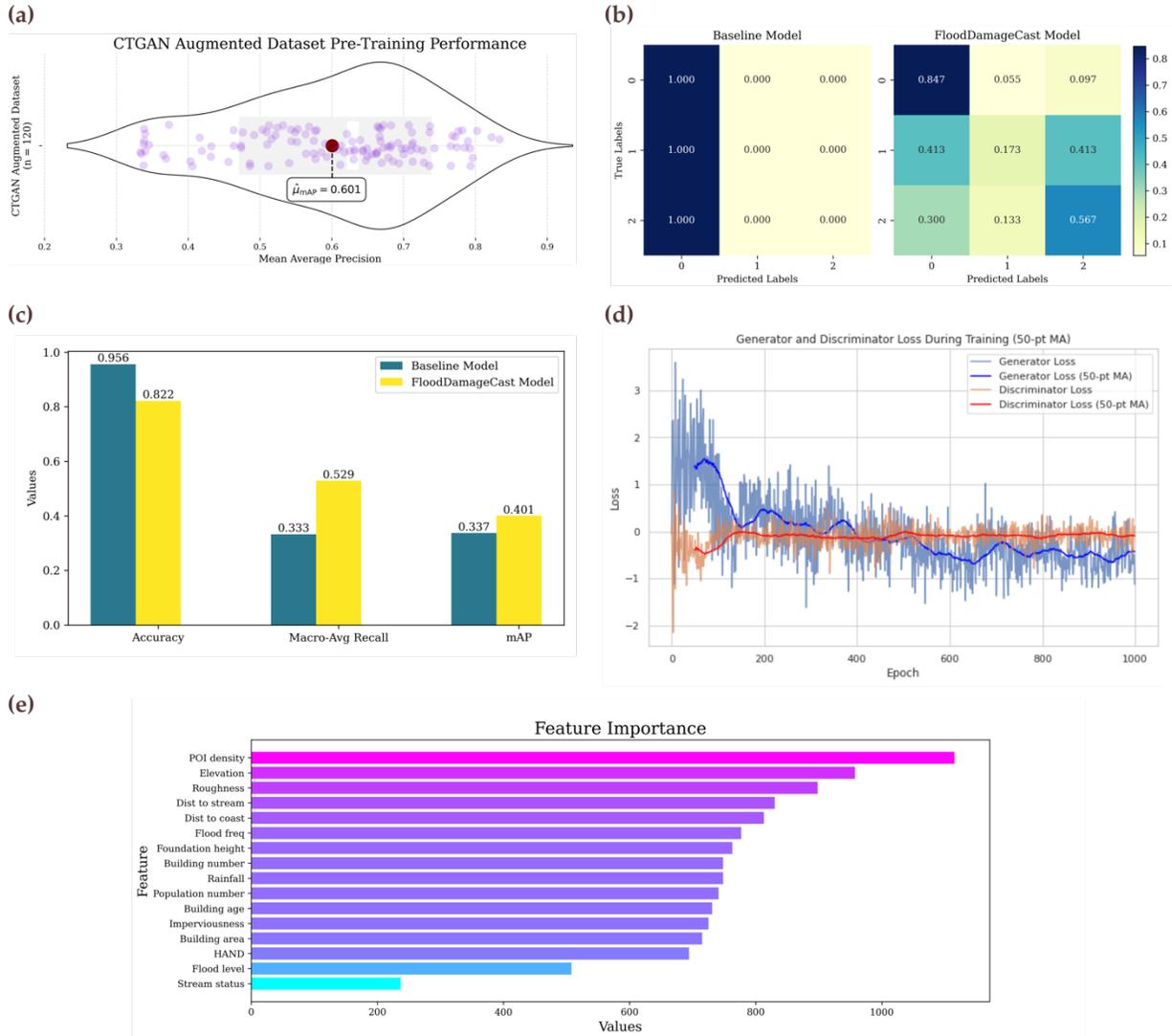

**Figure 5.** (a) Pretraining performance on CTGAN augmented dataset. (b) Normalize confusion matrices comparing the baseline model and FloodDamageCast model. (c) Comparative analysis of performance metrics: baseline model versus FloodDamageCast Model. (d) Generator and discriminator loss curves during CTGAN training with 0.001 generator learning rate and 0.001 discriminator learning rate. Each loss curve was attached to a 50-point moving average (MA) curve for smoother visualization. The curves indicate that the optimal stoppage epoch for both generator and discriminator is around 850. (e) FloodDamageCast feature importance. One of the built environment characteristics, POI density, serves as the most influential predictor of residential property damage. Four of the topographic characteristics, elevation, roughness, distance to stream, and distance to coast, show a higher feature importance relative to most built environment and hydrological characteristics.

## 5 Results

We present the model performance results related to different components of the FloodDamageCast model. The training loss for the CTGAN, as shown in Figure 5d, reveals that both the generator and discriminator converge with a learning rate of 1e-3 for each model. Initially, the generator's loss exhibits substantial fluctuations, which progressively diminish as training continues. Around epoch 850, which seems to be the optimal cessation point, both losses stabilize and show minimal variance, indicating effective convergence. By this stage, the discriminator's loss levels out, effectively distinguishing between real and generated data, while the generator's loss stabilizes, demonstrating enhanced capability in producing convincing synthetic data. Given the training regime's review at every 50 epochs up to a maximum of 1,000 epochs, epoch 850 emerges as a juncture where further training offers diminishing returns. Thus, the checkpoint saved at epoch 850 is advisable to capture the optimal model performance and prevent overfitting.

The optimal set of hyperparameters for the CTGAN model, identified via pretraining, achieves an average marginal distribution similarity of 90.21% across all attributes. To evaluate the performance of generated data for minority PDE classes, we compared the frequency distributions of real and synthetic attributes specific to these classes, as presented in Figure 6. Initially, PDE classes 1 and 2 comprise only 0.8% and 2.8% of all real data records, respectively, resulting in a ratio of approximately 1:3.5. In contrast, the synthetic data assigns an equal representation of 20% to both PDE class 1 and 2, altering the original ratio to an equal 1:1. Notably, the average marginal distribution similarity score across all attributes from these minority classes is 91.09%. This score surpasses the overall average, indicating that the CTGAN model has effectively prioritized minority classes during training, albeit at the expense of some performance loss on majority classes.

The heavily skewed distribution inherent in the damage claim dataset poses difficulties for conventional ML algorithms, particularly in accurately predicting instances belonging to minority classes, such as those representing high levels of residential building damage. As depicted in Figure 5b, the baseline model struggled to discern these minority class instances, instead classifying all examples as class 0. While achieving an overall accuracy of 0.956, the model failed to identify any instances from class 1 or class 2, which represent medium and high building damage, respectively, highlighting the necessity for a classifier capable of effectively distinguishing these critical instances without compromising accuracy for the majority class. To address this challenge, the FloodDamageCast model underwent the optimization of the imbalance ratio and LightGBM hyperparameter using the optimal augmented dataset. The result shows that the proposed model successfully classified 56.7% of high-damage (class 2) instances while keeping a satisfying accuracy of 0.822. It is noteworthy that the model could also identify the medium damage (class 1) instances but tends to overestimate the severity and classifies a significant number of them (0.413) as high damage. One potential explanation for this inferior performance in class 1 may lie in its inherent complexity and heterogeneity of the data distribution within class 1. In addition, the choice of clustering algorithm, such as K-means, and the selection of the number of clusters could also influence the model's performance.

To ensure the explainability of the FloodDamageCast model, we conducted a feature importance analysis based on split importance. Split importance measures the total number of times a features is used to split the data across all threes in the model. A higher split importance value indicates that the feature contains more information about the target variable when constructing DTs. As illustrated in Figure 5e, the analysis unveiled the prominence of POI density as the most influential built environment characteristic. POI density within each spatial area serves as a proxy for development density, thereby directly influencing the likelihood of building damage during a flood event. In contrast, features like building area and building number may not encapsulate these urban dynamics as effectively, potentially contributing to their lower feature importance scores. In addition, four of the topographic features—elevation, roughness, distance to stream, and distance to coast—emerged with higher feature importance relative to most built environment and hydrological characteristics. This prominence can be attributed to the direct influence of topographic features on the susceptibility of an area to flooding. Moreover, the spatial variability inherent in topographic features underscores their significance in predicting building damage during flooding, as they capture nuanced variations in flood risk zones within a region. These findings underscore the importance of considering both topographic and built environment characteristics in flood damage nowcasting models for enhanced accuracy and reliability.

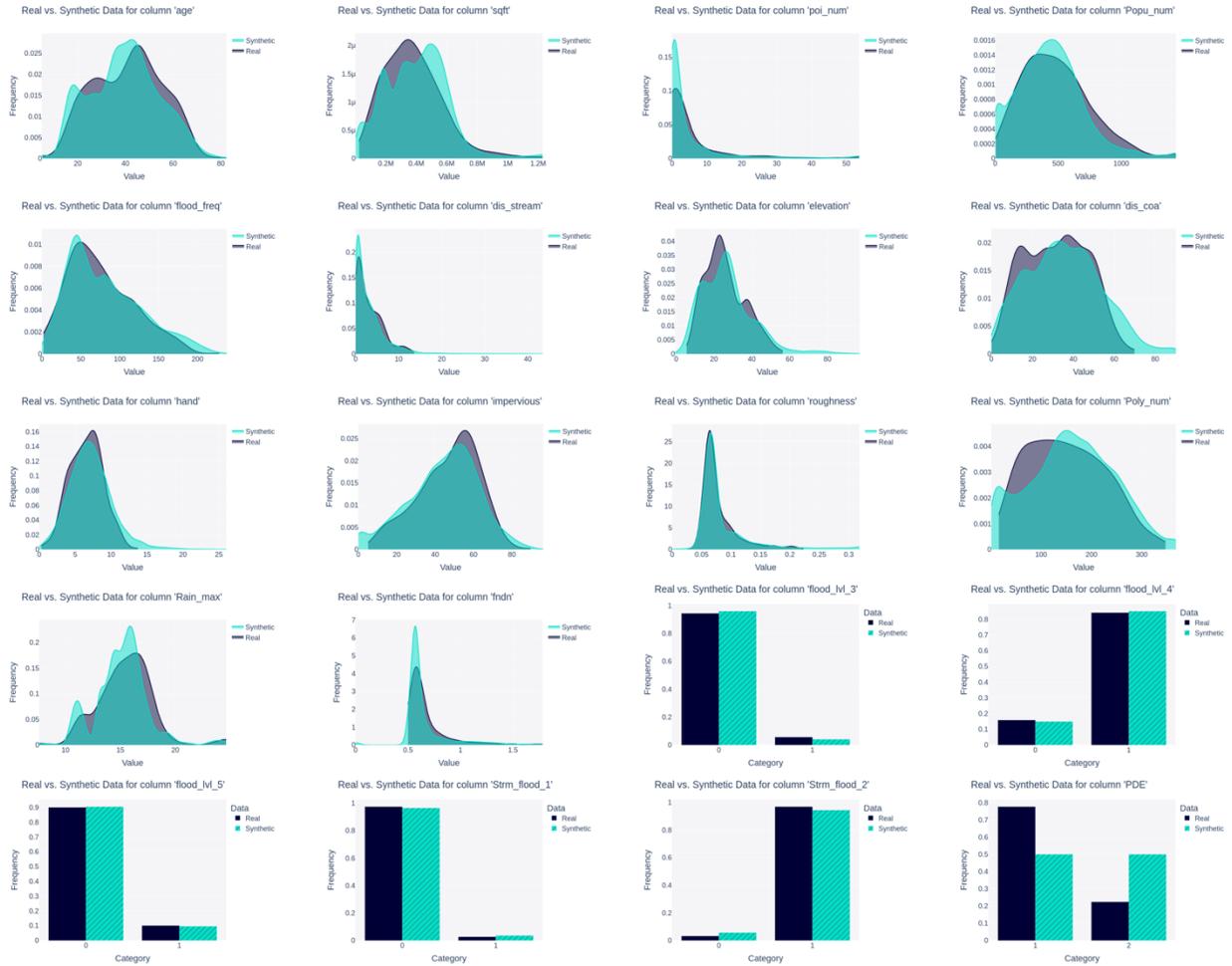

**Figure 6.** Stacked distribution comparison between real and synthetic attributes belonging to PDE class 1 and 2; the real distribution is shaded gray or black and the synthetic distribution is shaded blue. The subplot in the bottom left represents our target PDE categories, the synthetic classes ratio is controlled by the condition configured in CTGAN sampling. The rest of the subplots visualize a similar distribution between real and synthetic attributes.

## 6 Concluding remarks

Flood damage nowcasting is a critical step in flood response and recovery processes with a goal to provide rapid estimation of the extent of flood damage to buildings. Nowcasting of flood damage to residential buildings remains a challenging task with no antecedent in data-driven models and ML-based solutions. The main approach for flood damage estimation has been utilizing depth damage curves based on estimated inundation levels, work that is computationally expensive with significant uncertainty, making this approach not suitable for flood damage nowcasting [57]. To address this important gap, this study proposes the FloodDamageCast, a first-of-its-kind ML-based building flood damage nowcasting tool for rapid and automated assessment. Leveraging large empirical claim datasets, including NFIP and IA, this study proposes a data-driven methodology that explicitly accounts for various heterogeneous features related to the built environment, topographic, hydrologic, and event-specific hazard characteristics that shape flood damage extent across different spatial areas. The novelty of this study lies in creating an efficient ML model, LightGBM, using the CTGAN-augmented dataset. This data augmentation technique effectively addresses class imbalance issues commonly observed in the flood risk assessment by generating new data with a distribution similar to the target variable, thereby alleviating the data shortage problem in the minority class.

The proposed FloodDamageCast model encompasses multiple novel components. First, we use LightGBM as the main ML model for FloodDamageCast. LightGBM, renowned for its exceptional speed and memory efficiency, is a highly efficient ML algorithm. This enables us to conduct imbalance ratio learning on the augmented dataset and optimize model hyperparameters within a reasonable time frame. This combined approach heralds the development of the FloodDamageCast model, facilitating rapid damage assessment immediately following flood events. Second, FloodDamageCast predicts residential flood damage extent based on a heterogeneous set of input features obtained from publicly available datasets. Similar datasets exist in other regions and could be used in computing the features, and thus, FloodDamageCast could be adapted in other regions based on local datasets to enable rapid and automated flood damage nowcasting. Third, by leveraging the ML framework alongside GAN-based data augmentation techniques, the results demonstrate a significant enhancement in prediction performance, particularly in high-damage spatial areas. Fourth, FloodDamageCast has an explainability analysis that reveals important features contributing to flood damage extent. The results identify the POI density as a key indicator of urbanization or development density, as one of the most important features in shaping residential flood damage. In addition, four of the topographic features utilized in this study, including elevation, roughness, distance to stream, and distance to coast, emerge as highly important flood hazard variables contributing to residential building damage extent. It's worth noting that the two categorical predictors—flood level and stream status—both show lower split importance. This could be attributed to the typically lower information content in categorical variables, given their fewer distinct values compared to numerical variables. The explainability of FloodDamageCast model enables its use in post-event damage evaluations to inform future flood mitigation and risk reduction plans and measures.

From a practical perspective, FloodDamageCast can significantly enhance the ability of emergency managers and public officials to rapidly assess residential damage and thus, to expedite response and recovery processes. After flood hazards, emergency managers must expedite the damage evaluation process. A delay in on-site assessment could delay reimbursement of expenses for affected households, which may prolong the adverse impact of flooding on those urgently seeking to rebuild/repair and return to normalcy. Relying primarily on physical inspection for processing flood insurance claims could causes significant delays in processing claims, receipt of advance payments, and implementation of field adjustments. This delay adversely affects impacted residents' timely ability to repair their homes and significantly slows down the recovery process. This limitation is mainly due to: (1) inadequately trained personnel conducting field inspections in the aftermath of flood events; and (2) limited data-driven tools for rapid and automated damage nowcasting. Addressing these gaps, FloodDamageCast serves as a tool for rapid and automated estimation of flood damage at fine spatial resolutions, potentially expediting and supporting on-site residential building damage assessments. This tool enables local public officials and emergency managers to rapidly identify and prioritize the high-damage hotspots, facilitating the rebuilding and recovery of the highly-damaged areas.

While FloodDamageCast demonstrated a good performance for a the challenging task of flood damage nowcasting, additional steps could be undertaken for improvement in future work. In particular, it is important to note a discrepancy between the performance of the augmented data's validation set and the original data's testing set, which did not include data augmentation. The classification task in the pretraining yielded an mAP of 0.834, whereas the final learning outcome dropped to 0.401. Several factors may contribute to this divergence. First, there could be a distribution shift, suggesting that the augmented dataset may not fully capture the distribution of the original dataset. Consequently, significant disparity in data distributions between the augmented and original datasets may impede the model's ability to generalize from the augmented to the original data. Second, overfitting to the augmented data is a potential concern. This implies the LightGBM model may have memorized noise or specific patterns present in the augmented dataset during training, without learning the underlying generalizable patterns. Third, a lack of diversity in the augmented data may exist, leading to biases not present in the original dataset. Consequently, the model may struggle to generalize well to the broader range of examples in the original dataset. Addressing this discrepancy underscores the importance of refining the data augmentation process to ensure the augmented dataset accurately represents the original dataset. Although the FloodDamageCast has shown superior performance, compared with the baseline model, in identifying high-damage spatial areas, compared to the baseline model, there is still room to improve building damage assessment.

Some key recommendations for future research emerge from this study. Firstly, further exploration and refinement of data augmentation technique could improve the performance and generalization of the FloodDamageCast Model. This could involve investigating different augmentation methods or developing novel approaches to address distribution shifts and overfitting. Secondly, conducting a through analysis of feature engineering techniques and feature selection methods is recommended to enhance the predictive power of the nowcasting model. Exploring alternative predictors or incorporating domain-specific knowledge could improve the model's ability to capture relevant patterns and relationships. Lastly, conducting validation and benchmarking studies is helpful to evaluate the performance of the FloodDamageCast model against alternative approaches and existing flood damage estimation methods. Testing the model on diverse datasets from different regions and flood events will assess its generalization capability and robustness. Collectively, theses future research efforts can propel the field of flood damage nowcasting forward and foster more precise and effective decision-making in disaster response and recovery efforts.

**Acknowledgements:** This material is based in part upon work supported by the National Science Foundation under CRISP 2.0 Type 2 No. 1832662 and the Texas A&M University X-Grant 699. Any opinions, findings, conclusions, or recommendations expressed in this material are those of the authors and do not necessarily reflect the views of the National Science Foundation or Texas A&M University.

**Author contributions:** C.L. and A.M. designed the study framework. C.L., L.H. and A.M. wrote the main manuscript text. C.L. and L.H. prepared figures. All authors reviewed the manuscript.

**Competing interests:** The authors declare no competing interests.

**Code availability:** The code that supports the findings of this study is available from the corresponding author upon request.